\documentclass[letterpaper]{article} 
\usepackage{aaai2026}  
\usepackage{times}  
\usepackage{helvet}  
\usepackage{courier}  
\usepackage[hyphens]{url}  
\usepackage{graphicx} 
\usepackage{natbib}  
\usepackage{caption} 
\usepackage{algorithm}
\usepackage{algorithmic}
\usepackage{amsmath}
\usepackage{amssymb}
\usepackage{multirow} 
\usepackage{subcaption}
\usepackage{newfloat}
\usepackage{listings}
\DeclareCaptionStyle{ruled}{labelfont=normalfont,labelsep=colon,strut=off} 
\floatstyle{ruled}
\newfloat{listing}{tb}{lst}{}
\floatname{listing}{Listing}

\title{DANCE: Density-Agnostic and Class-Aware Network for Point Cloud Completion}
\author {
    Da-Yeong Kim\textsuperscript{\rm 1},
    Yeong-Jun Cho\textsuperscript{\rm 1}
}
\affiliations {
    \textsuperscript{\rm 1}Department of Artificial Intelligence Convergence\\
    Chonnam National University, Gwangju 61186, South Korea\\
    qetuo0909@jnu.ac.kr, yj.cho@jnu.ac.kr
}

\begin{document}

\maketitle

\begin{abstract}
      Point cloud completion aims to recover missing geometric structures from incomplete 3D scans, which often suffer from occlusions or limited sensor viewpoints.
      Existing methods typically assume fixed input/output densities or rely on image-based representations, making them less suitable for real-world scenarios with variable sparsity and limited supervision.
      In this paper, we introduce Density-agnostic and Class-aware Network (DANCE), a novel framework that completes only the missing regions while preserving the observed geometry.
      DANCE generates candidate points via ray-based sampling from multiple viewpoints. A transformer decoder then refines their positions and predicts opacity scores, which determine the validity of each point for inclusion in the final surface.
      To incorporate semantic guidance, DANCE includes a classification head and fusion network trained directly on geometric features, enabling category-consistent completion without relying on external image supervision.
      Extensive experiments on the PCN and MVP benchmarks show that DANCE outperforms state-of-the-art methods in accuracy and structural consistency, while remaining robust to varying input densities and noise levels.
\end{abstract}

\begin{links}
    \link{Code}{https://github.com/ayeong0909/DANCE}
\end{links}

\section{Introduction}

      Point cloud completion aims to recover missing geometric structures from incomplete 3D point clouds, typically caused by occlusions, limited sensor coverage, or constrained viewpoints.
    This task is essential for downstream applications like autonomous driving, robotics, and 3D reconstruction, where complete and accurate 3D representations are critical.
    The objective is to generate a plausible and structurally consistent point cloud that fills in missing regions while preserving the geometry of the observed input.

    Recent advances in deep learning have led to point cloud completion methods~\cite{yuan2018pcn} that generate complete shapes from partial inputs by learning global geometric priors.
    Building on this, recent generative approaches~\cite{li2025genpc, wei2025pcdreamer} leverage powerful image-to-3D models by converting partial point clouds into 2D images and performing shape completion using 2D generative priors.
    Meanwhile, partial completion methods~\cite{huang2020pf, yu2021pointr} focus on reconstructing only the missing regions, aiming to preserve the observed geometry and avoid regenerating known areas.

\begin{figure}
   \centering
   \begin{subfigure}[b]{0.48\columnwidth}
     \includegraphics[width=\linewidth]{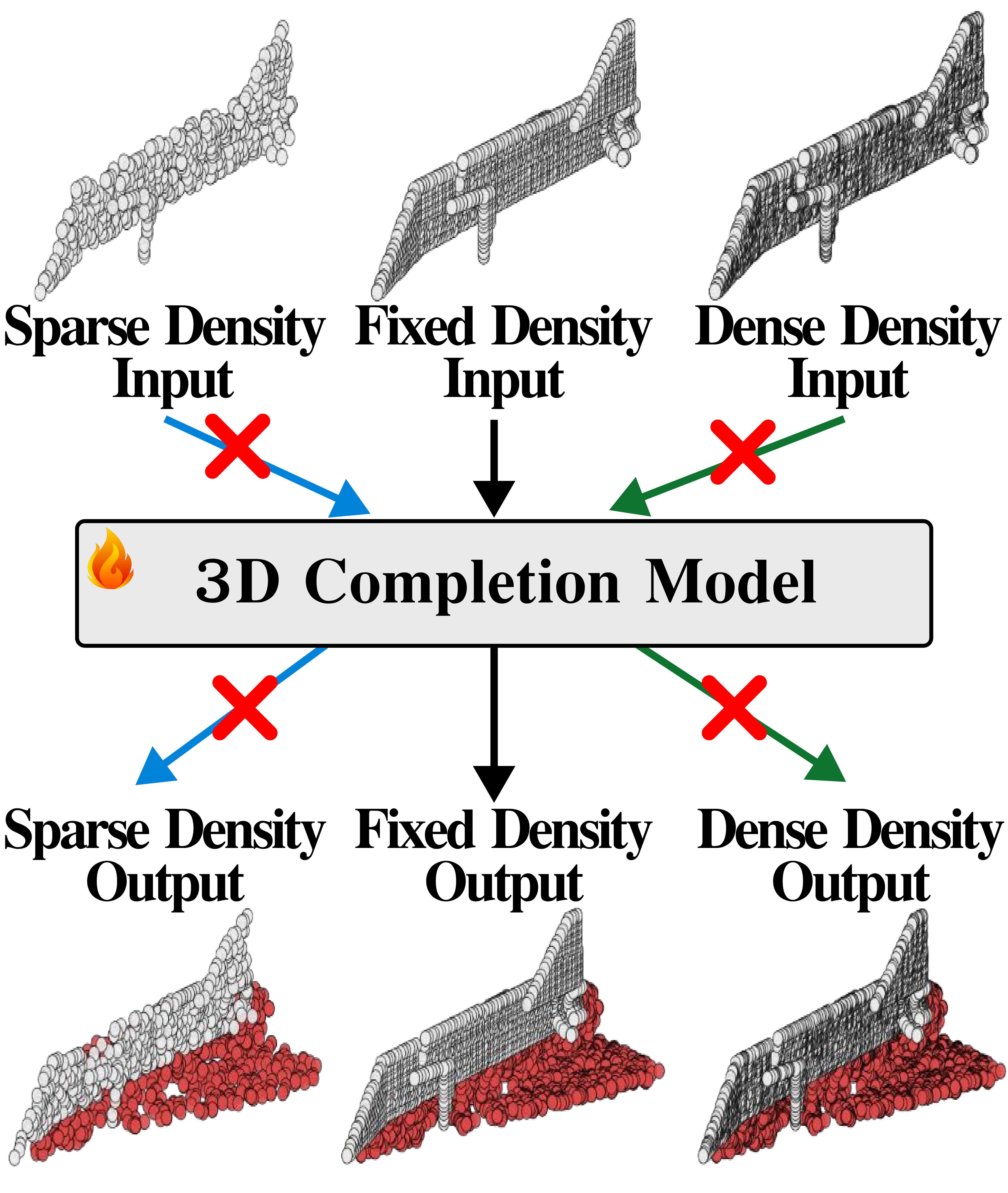}
     \caption{Fixed input and output}
   \end{subfigure}
   \begin{subfigure}[b]{0.48\columnwidth}
     \includegraphics[width=\linewidth]{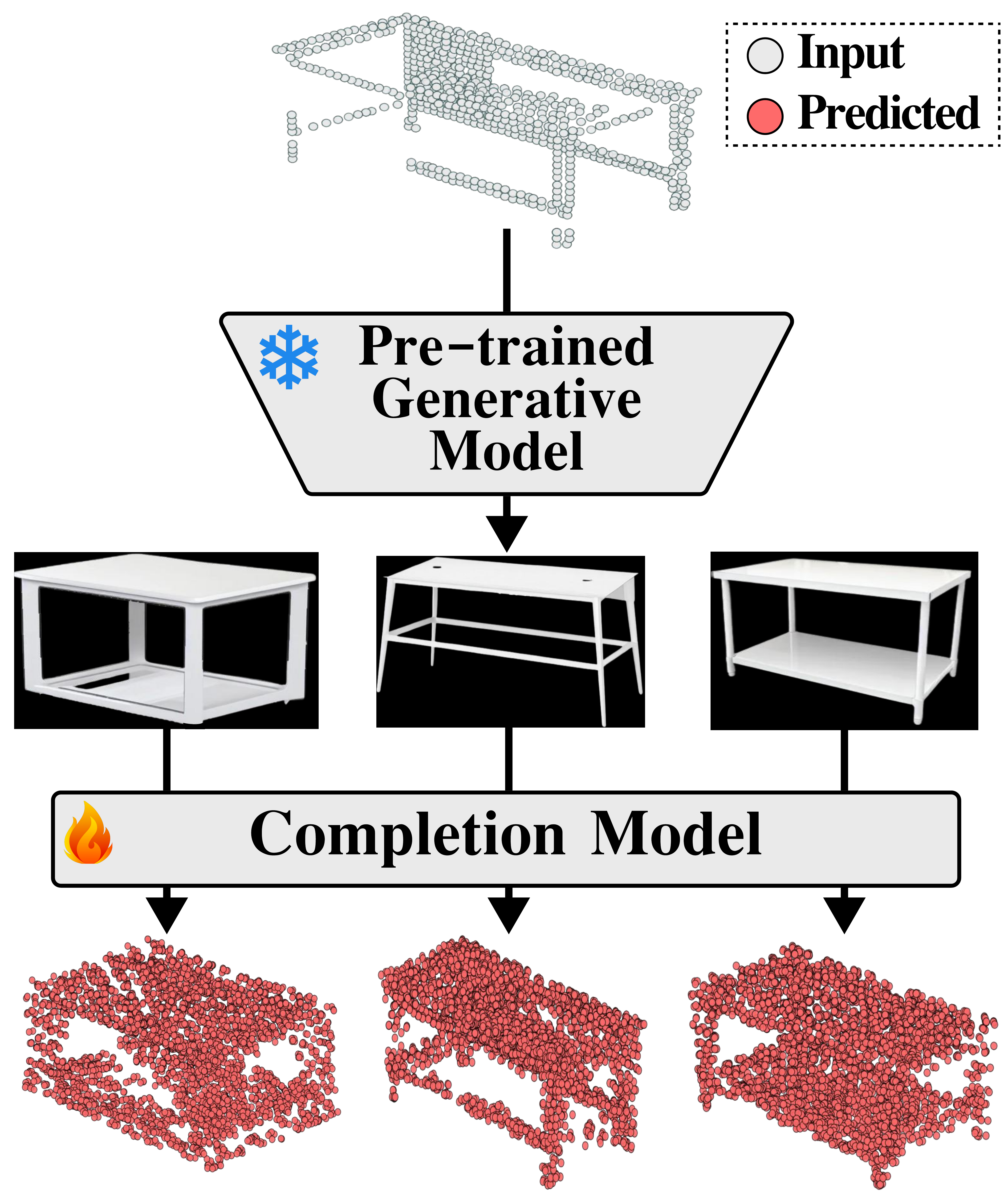}
     \caption{Relying on 2D supervision}
   \end{subfigure}
   \caption{Limitations of previous methods.
    (a) Fixed input/output densities. Prior methods assume fixed-size inputs and outputs, which limits their applicability in real-world scenarios.
    (b) Image-based generative frameworks. Methods relying on 2D image representations often produce shapes that do not align with the original 3D input geometry.}
   \label{fig:method_overview}
 \end{figure}

    Although these methods have achieved significant progress, they still suffer from two key limitations.
    First, most approaches assume fixed input and output densities, which is unsuitable for real-world scenarios (see Fig.\ref{fig:method_overview}(a)).
    In practice, input sparsity varies depending on object distance and sensor resolution, yet many methods generate a fixed number of points regardless of the actual level of incompleteness or the desired output density.
    Second, many methods incorporate semantic priors by relying on generative frameworks that use image-based representations.
    This reliance on external image representations prevents direct learning from 3D geometry.
    As shown in Fig.~\ref{fig:method_overview}(b), it often leads to structurally inconsistent outputs that deviate from the input object due to the strong bias of well-trained 2D generative models.

    To overcome these limitations, we propose Density-agnostic and Class-aware Network (DANCE) for point cloud completion.
    DANCE incorporates a lightweight classification head and fusion network trained directly on geometric features to address the lack of semantic guidance.
    Unlike prior methods that depend on image-based representations through generative frameworks, this module learns semantic priors purely from 3D point cloud data.
    In addition, DANCE adopts a density-agnostic design inspired by NeRF-style~\cite{mildenhall2021nerf} sampling.
    Given a partial point cloud, it generates candidate points via ray-based sampling over the observed surface and refines them using a transformer decoder.
    An opacity score is also predicted for each candidate to determine its inclusion in the output.
    This enables DANCE to process inputs of varying density and produce outputs without relying on fixed output density.

    We conduct extensive experiments on point cloud completion benchmarks (PCN, MVP) and demonstrate that DANCE outperforms state-of-the-art methods in both accuracy and structural consistency.
    Thanks to its density-agnostic design, our model maintains stable performance across varying input densities and generates outputs with flexible resolution.
    In addition, DANCE achieves superior class-aware completion quality, producing plausible and category-consistent shapes without relying on image supervision.
    Overall, DANCE provides a density-agnostic and class-aware completion framework that achieves flexible, semantically guided outputs without relying on image supervision.
    Our main contributions are as follows:
    \begin{itemize}
     \item We propose DANCE, a novel point cloud completion framework that is density-agnostic and class-aware, designed to overcome key limitations of existing methods.
     \item DANCE achieves state-of-the-art performance in accuracy and structural consistency on standard benchmarks such as PCN and MVP.
     \item Our method is scalable and practical for real-world scenarios, requiring no image supervision and robustly handling varying input sparsity.
    \end{itemize}
    To the best of our knowledge, this is the first attempt at density-agnostic 3D point cloud completion.

\section{Related Works}
 
    Thanks to recent progress in deep learning, many studies have focused on learning-based point cloud completion.
    Point cloud completion methods aim to generate complete shapes from partial inputs.  
    PCN~\cite{yuan2018pcn} introduced a pioneering encoder–decoder framework that extracts global features from partial point clouds and reconstructs complete shapes using a       FoldingNet~\cite{yang2017foldingnet}-based decoder.
    Recent studies have begun leveraging well-designed generative models, particularly 2D image-to-3D architectures, to improve point cloud completion.
    For example, GenPC~\cite{li2025genpc} converts partial point clouds into intermediate 2D image representations and applies pre-trained image-to-3D generative models to complete shapes.
    PCDreamer~\cite{wei2025pcdreamer} adopts a diffusion-based framework that gradually refines 3D shapes by learning latent distributions. 
    It combines geometric cues and learned shape priors to generate high-quality completions, even in the presence of large missing regions.

    Although generative model-based completion methods demonstrate impressive completion performance, they still have several limitations.
    First, regenerating the entire point cloud, including visible regions, can lead to geometric distortions and the loss of fine details originally present in the input.
    Second, methods that require paired image–point cloud datasets are impractical in real-world settings due to the high cost and difficulty of data collection.

    While many recent methods adopt generative models that regenerate the entire point cloud, some earlier approaches have focused on completing the missing regions to preserve the observed geometry.
    For instance, PF-Net~\cite{huang2020pf} introduces a pyramid decoder that generates missing points at multiple scales while retaining the original input points throughout the process.
    PoinTr~\cite{yu2021pointr} applies transformers to point cloud completion by dividing the input into small groups called ``point proxies'' that represent local regions.
    Recently, although not limited to missing region completion, transformer-based methods such as SeedFormer~\cite{zhou2022seedformer} and SVDFormer~\cite{zhu2023svdformer} combined partial completion with global shape modeling.
    
    Despite their strengths, these completion methods still have important limitations.
    One key issue is their fixed input and output density, which is not suitable for real-world scenarios.
    In practice, input point clouds can be either sparse or dense, depending on factors such as the distance to the object or the resolution of the 3D sensor.
    Moreover, the appropriate output density is typically undefined.
    For example, they generate a fixed number of points regardless of how much of the shape is missing, resulting in unrealistic and uneven point distributions.
    Another limitation is that these methods generally lack semantic awareness and treat all object categories uniformly, without leveraging category-specific shape priors that could guide more accurate completion.

    \begin{figure*}[t]
      \centering
      \includegraphics[width=1\textwidth]{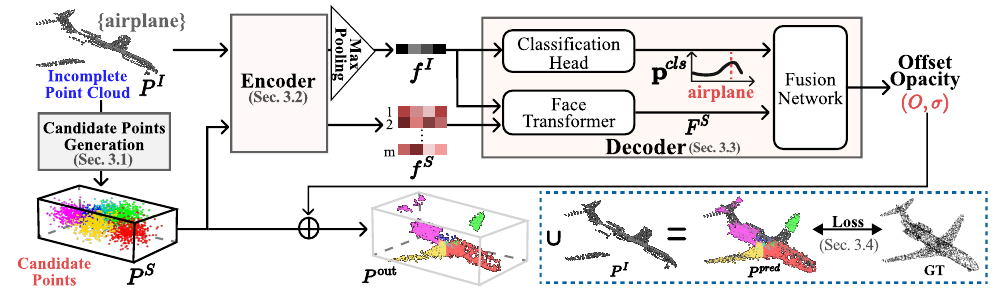}
      \caption{The overall pipeline of DANCE. It generates candidate points using ray-based sampling strategy to reconstruct missing regions of incomplete point clouds. The encoder extracts features from both the incomplete input and generated points, and the decoder processes them to estimate the offset and opacity of each point. Final completion results are obtained by refining candidate points using the predicted offset and selecting valid ones based on the predicted opacity.
      Best viewed in color.}
      \label{fig:main_figure}
   \end{figure*}

\section{Proposed Methods}
   The goal of point cloud completion is to reconstruct the complete 3D point cloud of an object by predicting the missing geometric structures from an incomplete point cloud $P^I\in \mathbb{R}^{N \times 3}$, where $N$ is the number of points.
   The proposed DANCE first generates candidate points to predict missing regions in 3D space (Sec.~3.1).
   The candidate points and given incomplete point cloud are processed through an encoder-decoder architecture.
   The encoder (Sec.~3.2) extracts 3D features from both candidate points and given incomplete point cloud, while the decoder (Sec.~3.3) reconstructs the geometry of the missing regions $P^{\text{out}}$.
   The overall pipeline of the proposed DANCE is illustrated in Fig.~\ref{fig:main_figure}.

    Most existing point cloud completion methods~\cite{li2025genpc, wei2025pcdreamer} generate outputs with a fixed density, regardless of the input’s sparsity or density.
    In contrast, the proposed DANCE is density-agnostic: it works with inputs of varying densities and does not require a fixed output size.
    Additionally, DANCE leverages the classification results of the incomplete point cloud $P^I$ to guide the completion process, allowing semantic information to improve completion accuracy.
    Thus, DANCE performs density-agnostic and class-aware point cloud completion, enabling significantly more natural and accurate reconstruction results, and making it well-suited for real-world applications with varying input conditions.

   \subsection{Candidate Points Generation}
   \label{subsec:point_generation}   
   Motivated by NeRF~\cite{mildenhall2021nerf}, which samples 3D points along multi-view camera rays for novel view synthesis, we adopt a similar ray-based sampling strategy for point cloud completion.
   NeRF predicts color and opacity at sampled 3D points to synthesize novel views, as its goal is image rendering rather than 3D reconstruction.
   In contrast, our task focuses on completing missing 3D geometry from incomplete point clouds. Therefore, instead of predicting color as in NeRF, we focus on estimating the 3D positions and opacity of the candidate points.
    \begin{figure}[t]
      \centering
      \includegraphics[width=0.85\linewidth]{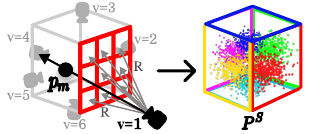}
      \caption{Candidate point generation in DANCE. Candidate points are generated using a ray-based sampling strategy. Colors represent the corresponding faces and their associated candidate points. Best viewed in color.}
      \label{fig:overview_sampling}
   \end{figure}

   To reconstruct the missing regions of an incomplete given point cloud $P^I$, we generate a set of candidate points $P^S$.  
    To this end, we define $V$ viewpoints surrounding $P^I$, each associated with a face oriented toward the object, as illustrated in Fig.~\ref{fig:overview_sampling}.
    For example, setting $V=6$ allows the six viewpoints and faces to form a hexahedral structure that surrounds the object.
    An $R \times R$ grid is placed on each face, and rays are cast from the corresponding viewpoint through each grid location.
    A single 3D point is then sampled along each ray based on a Gaussian distribution centered between the face and the object surface.
    As object point clouds are defined only on the surface, sampling one point per ray is sufficient for representing the geometry.
    This results in the initial candidate point set $P^S = \{p_m\}^{M}_{m=1} \in \mathbb{R}^{M \times 3}$, where $M = VR^2$ denotes the total number of the points.
    Here, $P^S_v \in \mathbb{R}^{R^2 \times 3}$ denotes the set of candidate points corresponding to the $v$-th viewpoint.
    Each $P^S_v$ is visualized with a different color in Fig.~\ref{fig:overview_sampling}.
    Note that $P^S$ is not perfect yet, but serves as the starting point for reconstruction.  
    It is subsequently refined into accurate 3D positions through the encoder-decoder network.
   
   \subsection{Encoder for 3D Feature Extraction}
       \begin{figure}[t]
      \centering
      \includegraphics[width=0.85\linewidth]{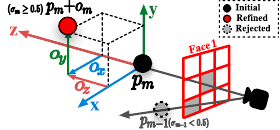}
      \caption{Offset and opacity estimation in DANCE. The estimated offset and opacity refine the position of each candidate point and control its inclusion in the final completion.
      }
      \label{fig:overview_offset}
   \end{figure}
   
    A set of candidate points $P^{S}$ and the incomplete point cloud $P^{I} $ are fed into the 3D encoder $E$ to extract candidate and global 3D features as follows:
    \begin{equation}
      f^{S}= E\left(P^{S}\right) \in \mathbb{R}^{M\times d_{\text{en}}},
   \end{equation}
   \begin{equation}
   \label{eq:2}
      f^{I}= \text{maxpool}\left(E(P^{I})\right) \in \mathbb{R}^{1\times d_{\text{en}}},
   \end{equation}
   where $d_{\text{en}}$ denotes the dimension of the feature vector.  
    Note that $P^I$ additionally passes through a max-pooling layer, as defined in Eq.~\ref{eq:2}, to extract a global feature. 
    Although the global feature $f^I$ is derived from the incomplete point cloud $P^I$, it serves as a compact latent representation that encodes global shape priors of the visible regions.  
    In parallel, the candidate feature $f^{S}$, extracted from the candidate points $P^{S}$, contributes to the prediction of the missing structures.

    As the backbone for the encoder, any architecture such as the PointNet series~\cite{qi2017pointnet, qi2017pointnet++} or DGCNN~\cite{wang2019dynamic} can be used.  
    $P^{S}$ and $P^{I}$ share the same encoder $E$,  
    which not only improves efficiency but also ensures that both the candidate points and the incomplete point cloud are represented in a consistent feature space.

   \subsection{Decoder for Point Cloud Completion}
   \label{subsec:decoder}

    The decoder consists of three main components:  
    1) a face transformer; 2) a classification head; and 3) a fusion network.  
    It performs point cloud completion by predicting the 3D coordinate offset and opacity value for each candidate point in $P^S$.
    Based on these predictions, the missing 3D point cloud $P^{\text{out}}$ is estimated.  
    Importantly, the proposed decoder focuses solely on reconstructing the missing regions, rather than regenerating the entire object point cloud.
    Moreover, thanks to its transformer-based design, the decoder can handle input point clouds with varying densities.

    We define the offset $o_m = \{o_x, o_y, o_z\}$ to represent the $(x, y, z)$ displacements of the $m$-th candidate point $p_m$.
    For each candidate point $p_m$, a local coordinate system is defined with $p_m$ as the origin, where the sampling ray is aligned with the $z$-axis, and the two axes of the corresponding face define the $x$- and $y$-axes, as shown in Fig.~\ref{fig:overview_offset}.
    The opacity value $\sigma$ represents the influence of each point, similar to its role in NeRF~\cite{mildenhall2021nerf}.
    Only points with $\sigma$ above a certain value ($\geq0.5$) are considered meaningful, while the rest are filtered out.
    This decoder design allows the model to adjust the output point cloud density.
    Then, the missing 3D point cloud $P^{\text{out}}$ is defined as follows:
    \begin{equation}
    \label{eq:p_out}
      P^{\text{out}} = \{ p_m + o_m \mid \sigma_m \geq 0.5,\; m = 1, \ldots, M \}.
   \end{equation}
   The final completion result is obtained by combining the predicted points $P^{\text{out}}$ with the incomplete input $P^I$, yielding $P^{\text{pred}} = P^I \cup P^{\text{out}}$.

   \paragraph{3.3.1. Face Transformer.}
    \begin{figure}[t]
      \centering
      \includegraphics[width=1\linewidth]{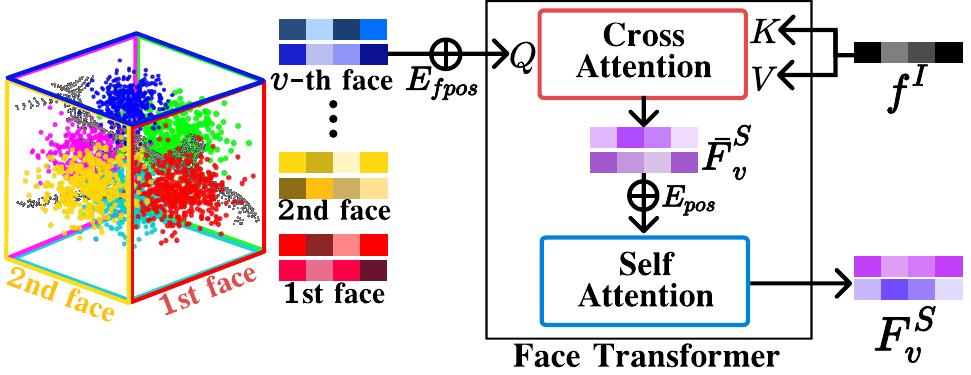}
      \caption{Architecture of the transformer in DANCE. Cross-attention between $f^S_v$ and $f^I$ followed by self-attention within viewpoint groups $v$.}
      \vspace{-5pt}
      \label{fig:transformer_architecture}
   \end{figure}
   
    The face transformer enhances the interaction between the candidate feature $f^S$ and the global feature $f^I$ through cross- and self-attention layers as shown in Fig.~\ref{fig:transformer_architecture}.  
    We assume that points generated from the same viewpoint share similar geometric and visibility characteristics.
    To reflect this, we process the candidate feature set $f^{S}_{v}$ from each viewpoint $v$ independently.
    Each candidate feature $f^{S}_{v}$ undergoes cross-attention with the global feature $f^I$ as follows: \begin{equation}
    \bar F^{S}_{v}= \text{crossAttn}(Q=f^{S}_{v}+ E^{fpos}_{v}, K=f^{I}, V=f^{I}),
    \label{eq:cross_attention}
    \end{equation}
    where $E^{fpos}_{v}$ is the positional embedding of viewpoint, maintaining the spatial relationships between different viewpoints.
    Since $f^I$ is a compact latent representation encoding global shape priors from the visible regions, this cross-attention allows each candidate point to incorporate global structural context.  
    As a result, the point features are refined to better guide the completion of missing regions.
   
   To further enhance local geometric consistency, we apply self-attention within each $\bar{F}^{S}_{v}$ as follows:
   \begin{equation}
      F^{S}_{v}= \text{selfAttn}(\bar F^{S}_{v}+E_{pos}), 
      \label{eq:self_attention}
   \end{equation}
   where $Q,K$ and $V$ denote the query, key, and value in the transformer. $E_{pos}$ is a positional embedding for each candidate point.
   Points generated from the same viewpoint exhibit strong spatial correlation; therefore, this intra-group self-attention helps enforce consistency among neighboring points.
   After applying self-attention to each set of features $\bar F^{S}_{v}$, the whole feature representation $F^{S}=\{ F^{S}_{1}, F^{S}_{2},\ldots, F^{S}_{V}\}$ is obtained. 
   The proposed two-stage attention simultaneously achieves both global consistency and local precision for reconstructing missing regions.
   
   \paragraph{3.3.2. Classification Head.}
   Most existing studies rely solely on spatial structure of the object for point cloud completion.
    However, even from incomplete point clouds $P^I$, it is possible to infer the object’s class distribution.
    In the proposed DANCE, this semantic information is utilized as a strong prior to guide and improve the reconstruction of missing regions.
    The classification head is implemented using multi-layer perceptrons (MLP) followed by a softmax function:
    \begin{equation}
      \mathbf{p}^{cls}=\text{softmax}\left(\text{MLP}(f^{I})\right)\in \mathbb{R}^{c},
      \label{eq:classification}
   \end{equation}
   where $c$ is the number of categories, and $\mathbf{p}^{cls}$ represents the class probability distribution.
   
   \begin{figure}[t]
      \centering
      \includegraphics[width=1\linewidth]{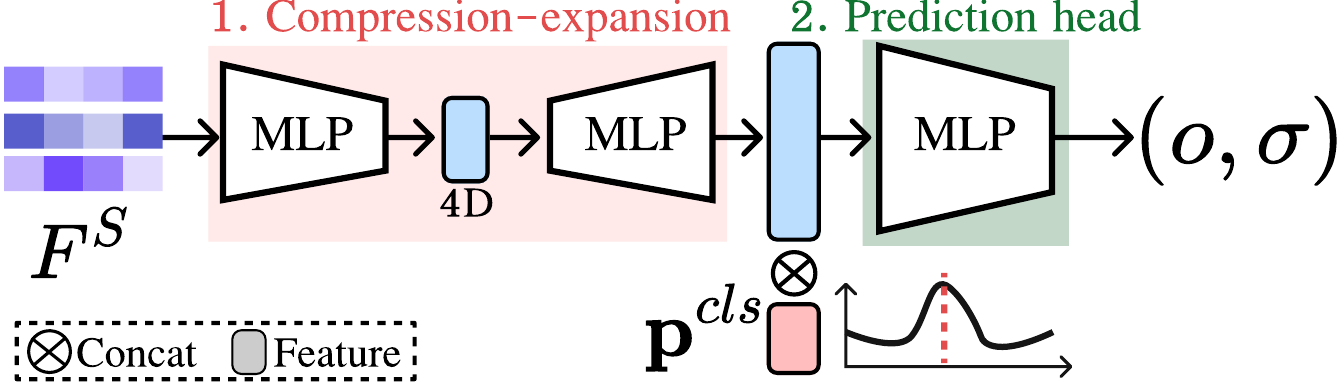}
      \caption{The architecture of the fusion network for offset and opacity prediction. The geometric feature $F^S$ is refined by incorporating classification results $\mathbf{p}^{cls}$ to predict $o,\sigma$.}
      \vspace{-5pt}
      \label{fig:mlp_architecture}
   \end{figure}

   \paragraph{3.3.3. Fusion Network}
   Finally, the fusion network integrates point feature representation $F^S$ with the predicted class distribution $\mathbf{p}^{cls}$ to estimate the final offsets and opacities of the candidate points $P^S$.
   The fusion network consists of a compression-expansion MLP and a prediction head as shown in Fig.~\ref{fig:mlp_architecture}.
   First, the compression-expansion MLP compresses the input feature representation $F^S\in \mathbb{R}^{  M\times d_{\text{en}}}$ to a 4-dimensional vector and then reconstructs it back to the original dimension.
   The 4D bottleneck is designed to match the dimensionality of the prediction head’s output (offset and opacity), encouraging the model to focus on essential task-specific cues.

   Next, the output vector of the MLP is concatenated with the class probability distribution $\mathbf{p}^{cls}$.
   The combined representation is then passed through the prediction head to infer the offset and opacity for each candidate point.
   Finally, we obtain the offset $o_m = \{o_x, o_y, o_z\}$ and the opacity value $\sigma_m$ for each candidate point $p_m$.
   Based on the predicted offsets and opacity values, the missing 3D point cloud $P^{\text{out}}$ is computed as described in Eq.~\ref{eq:p_out}.

   \subsection{Loss Functions}
   To train the proposed method, we jointly optimize the completion loss and the classification loss.
    The completion loss measures the geometric difference between the predicted point cloud $P^{pred}$ and the ground truth point cloud $P^{GT}$. To this end, we employ Chamfer Distance (CD), defined as:
   \begin{equation}
      \mathcal{L}_{CD}= \text{CD}(P^{pred}, P^{GT}).
      \label{eq:Chamfer_distance}
   \end{equation}
   It measures the similarity between two point clouds by summing the distances from each point in one set to its nearest neighbor in the other set.

   The classification loss encourages the model to correctly predict the object category from the incomplete point cloud feature $f^I$.  
    We compute the cross-entropy loss between the predicted class probability distribution $\mathbf{p}^{cls}$ and the ground truth one-hot label vector $\mathbf{y} = \{y_1, ..., y_c\}$ as follows:
   \begin{equation}
    \mathcal{L}_{cls} = - \sum_{i=1}^{c} y_i \log(\mathbf{p}_i^{cls}).
      \label{eq:feature_ext}
   \end{equation}
   The classification loss provides shape priors that guide the network to complete partially missing regions more reliably.
   The total loss function is formulated as a weighted sum of the completion and classification losses:
   \begin{equation}
      \mathcal{L}_{total}= \lambda\mathcal{L}_{CD}+ (1-\lambda) \mathcal{L}_{cls}.
      \label{eq:feature_ext}
   \end{equation}
   where $\lambda$ is a hyperparameter that balances the influence of the classification loss.

   \section{Experimental Results}
      \subsection{Datasets and Settings}
   \noindent
    \noindent\textbf{Datasets.} \quad We evaluate our method on two widely used point cloud completion benchmarks: PCN~\cite{yuan2018pcn} and MVP~\cite{pan2021variational}, both constructed from subsets of ShapeNet~\cite{chang2015shapenet}.
    PCN includes 8 object categories (e.g., Plane, Chair, Table) and provides approximately 30,000 complete point clouds and 240,000 incomplete ones. 
    The incomplete point clouds are generated by removing parts from 8 different viewpoints, with 2,048 points per sample, while complete shapes consist of 4,096 points.
    MVP, on the other hand, covers 16 categories and introduces more diverse partiality by simulating 26 viewpoints. 
    Although the input size remains fixed at 2,048 points, it offers complete ground truths at multiple resolutions (4,096, 8,192, and 16,384 points).

   \noindent\textbf{Settings.} \quad All experiments were conducted with ground truth point clouds containing 4,096 points.
   We set V=6 and R=21 for ray-based candidate points generation on each face.
   The opacity threshold was fixed at 0.5 to filter candidate points based on their predicted opacity values.
   We employ Chamfer Distance (CD), Density-aware CD (DCD), and F1-Score@1\% as evaluation metrics. CD-avg measures the average nearest-neighbor distance between predicted and ground truth point clouds, while DCD-avg improves upon CD by taking local point density into account. F1-Score@1\% is defined as the harmonic mean of precision and recall at a 1\% distance threshold. Following prior works, we use L1-based CD for the PCN and L2-based CD.

   \begin{figure}[t]
      \centering
      \includegraphics[width=1\linewidth]{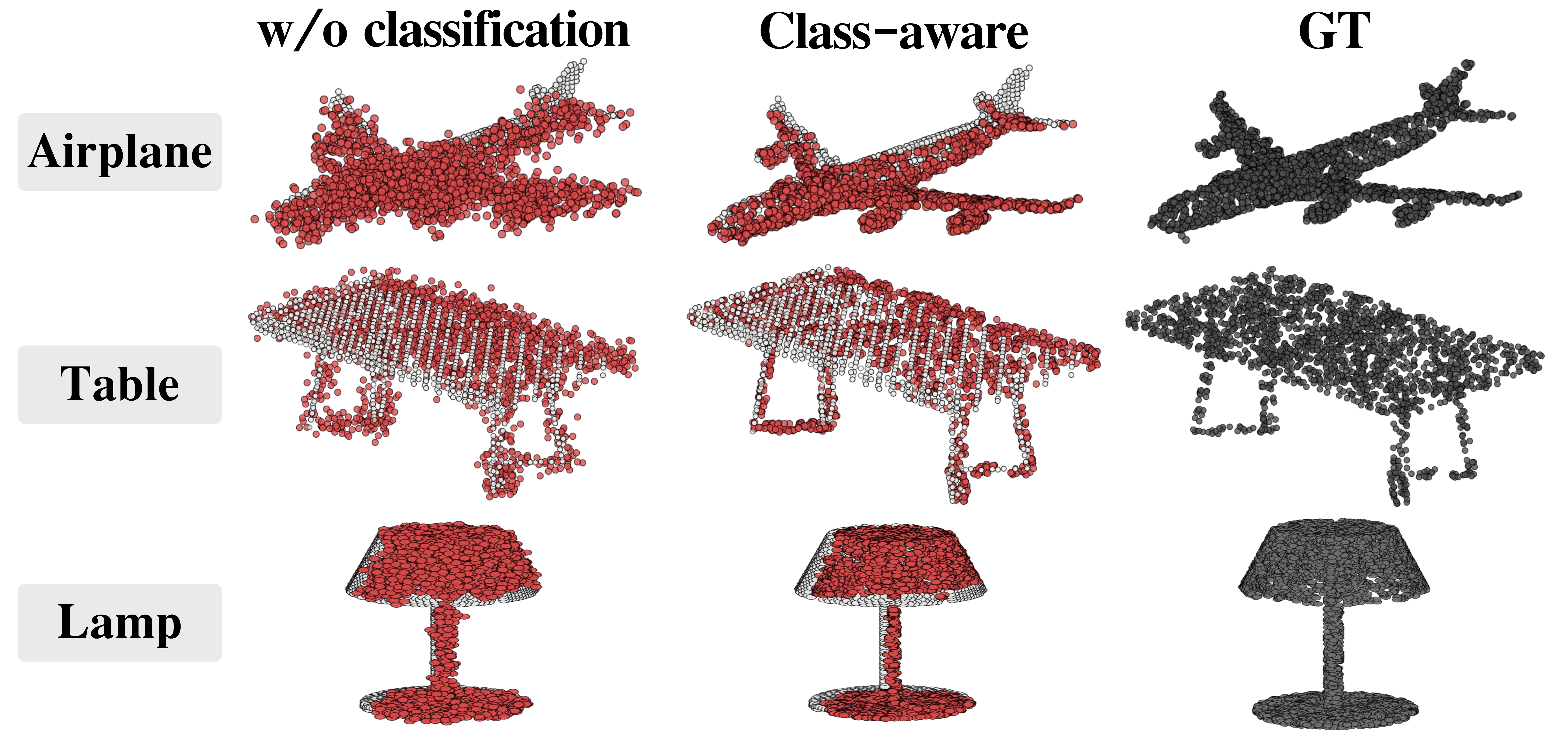}
      \caption{Effects of the proposed class-aware approach. Completion results with and without classification head. Gray and red points represent the input and predicted completions, respectively.}
      \label{fig:classification_effects}
   \end{figure}

   \begin{table}[t]
      \centering
      \renewcommand{\arraystretch}{0.95}
      \begin{tabular}{c|ccc}
         \hline
                Methods    & CD-Avg$\downarrow$ & DCD-Avg$\downarrow$ & F1$\uparrow$   \\
         \hline
         w/o classification & 6.46               & 0.528               & 0.856          \\
         w/o face attention & 6.52               & 0.534               & 0.849         \\
         Ours    & \textbf{6.42}      & \textbf{0.524}      & \textbf{0.859} \\
         \hline
      \end{tabular}
      \caption{Ablation study of proposed methods. Effectiveness of each component measured on PCN dataset. The full model achieves the best performance across all evaluation metrics.}
      \vspace{-5pt}
      \label{tab:effect_proposed}
   \end{table}

   \begin{figure*}[t]
      \centering
      \includegraphics[width=1\linewidth]{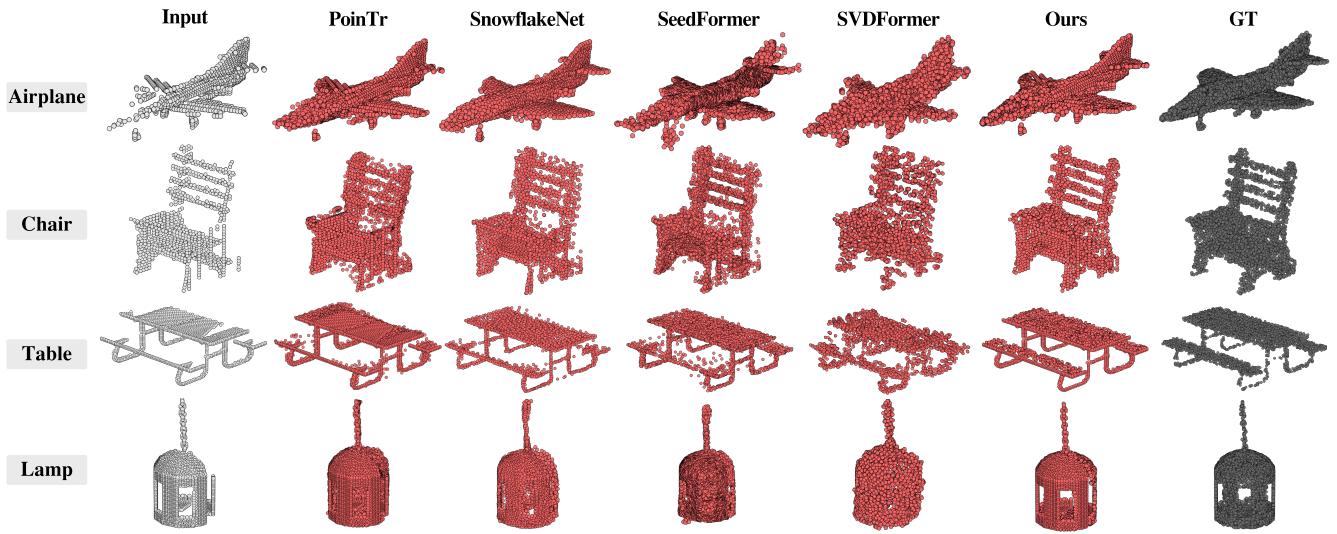}
      \caption{Qualitative Comparison with state-of-the-art methods on the PCN dataset.
      DANCE generates more complete and structurally consistent shapes compared to prior methods. 
      The completions better preserve the observed geometry while producing plausible details, demonstrating the effectiveness of our density-agnostic and class-aware design.
      }
      \label{fig:visualize_overall}
   \end{figure*}

   \begin{table*}
      [t]
      \small
      \centering
      \begin{tabular}{l|cccccccc|c|c|c}
         \hline
         Methods      & Plane         & Cabinet      & Car           & Chair         & Lamp          & Couch         & Table         & Boat          & CD-Avg$\downarrow$ & DCD-Avg$\downarrow$ & F1$\uparrow$   \\
         \hline
         PCN~\shortcite{yuan2018pcn}          & 5.82          & 10.91        & 9.00          & 11.09         & 11.91         & 12.06         & 9.56          & 9.12          & 9.93               & --               & 0.657          \\
         GRNet~\shortcite{xie2020grnet}        & 8.51          & 12.84        & 10.17         & 12.13         & 11.83         & 12.64         & 11.16         & 9.84          & 11.14              & 0.765               & 0.563          \\
         PointTr~\shortcite{yu2021pointr}      & 4.31          & 9.23         & 7.60          & 8.35          & 8.27          & 9.38          & 7.97          & 7.01          & 7.76               & 0.562               & 0.810          \\
         PMP-Net++~\shortcite{wen2022pmp}    & 4.80          & 11.70        & 8.96          & 8.93          & 7.04          & 12.25         & 8.57          & 6.88          & 8.64               & 0.714               & 0.683          \\
         SnowFlakeNet~\shortcite{xiang2021snowflakenet} & 3.95          & 8.82         & 7.52          & 7.48          & 6.34          & 8.88          & 6.61          & 6.10          & 6.96               & 0.567               & 0.828          \\
         SeedFormer~\shortcite{zhou2022seedformer}   & 3.87          & 9.05         & 7.53          & 7.63          & 6.12          & 9.26          & 6.61          & 6.03          & 7.01               & 0.573               & 0.824          \\
         AnchorFormer~\shortcite{chen2023anchorformer} & 3.62          & 8.79         & 7.20          & 7.12          & 6.18          & 8.71          & 6.50          & 6.01          & 6.77               & 0.539               & 0.841          \\
         HyperCD~\shortcite{lin2023hyperbolic}      & 3.87          & 9.05         & 7.40          & 7.50          & 5.95          & 9.25          & 6.50          & 5.89          & 6.93               & 0.568               & 0.822          \\
         SVDFormer~\shortcite{zhu2023svdformer}    & 3.68          & 8.73         & 7.10          & 6.95          & 5.64          & 8.58          & 6.26          & 5.91          & 6.61               & 0.534               & 0.848          \\
         CRA-PCN~\shortcite{rong2024cra}      & 3.62          & 8.77         & 7.00          & 6.92          & 5.46          & 8.59          & 6.27          & 5.86          & 6.56               & 0.537               & 0.846          \\
         PCDreamer~\shortcite{wei2025pcdreamer}    & \textbf{3.51} & 8.62         & 6.92          & 6.91          & 5.66          & 8.31          & 6.27          & 5.90          & 6.52               & 0.531               & 0.856          \\
         \hline
         Ours         & 3.53          & \textbf{8.6} & \textbf{6.88} & \textbf{6.89} & \textbf{5.44} & \textbf{8.28} & \textbf{6.24} & \textbf{5.82} & \textbf{6.46}      & \textbf{0.528}      & \textbf{0.859} \\
         \hline
      \end{tabular}
      \caption{Performance comparison with state-of-the-art methods on the PCN dataset.
      DANCE achieves the best overall performance across all metrics, consistently outperforming existing methods. The best results are highlighted in bold.}
      \vspace{-5pt}
      \label{tab:pcn_comparison}
   \end{table*}
   \subsection{Effectiveness of the Proposed Methods}
   To evaluate the individual contributions of our proposed methods, we conducted comprehensive ablation studies on the PCN dataset~\cite{yuan2018pcn}.
   We first evaluated the effectiveness of class-aware completion using the proposed decoder structure introduced in Sec.3.3.
   As shown in Tab.\ref{tab:effect_proposed}, the model with the classification head outperforms the baseline without the head across all evaluation metrics.
   Notably, we observed a reduction in CD-Avg from 6.46 to 6.42, indicating improved  reconstruction accuracy.
   The results in Fig.~\ref{fig:classification_effects} also demonstrate the effectiveness of the class-aware approach. 
   It produces more geometrically coherent completions and better preserves category-specific structural characteristics.
   This result supports that semantic priors extracted from incomplete inputs help the network resolve ambiguities in missing regions.

   We evaluated the effectiveness of the face transformer in the decoder, which performs self-attention within viewpoint-specific groups, by comparing models with and without this design. As shown in Tab.~\ref{tab:effect_proposed}, removing this component consistently degraded performance across all metrics.
   The results highlight that performing self-attention within viewpoint-specific groups enhances completion accuracy by exploiting local geometric coherence.
 
    \subsection{Comparison with State-of-the-Art Methods}
    We compared DANCE against state-of-the-art point cloud completion methods on the PCN and MVP benchmarks.
    As shown in Tab.~\ref{tab:pcn_comparison}, our method achieves superior performance across most object categories, with an overall CD-Avg of $6.46$ on the PCN dataset.
    Figure~\ref{fig:visualize_overall} presents qualitative comparisons with other point cloud completion methods.
    Our approach achieves more accurate reconstruction of fine geometric details while maintaining global shape consistency.
    For objects with complex structures, such as lamps and chairs, DANCE effectively recovers thin parts and symmetric components that existing methods often fail to reconstruct.
    
\begin{figure}[t]
      \centering
      \includegraphics[width=0.9\linewidth]{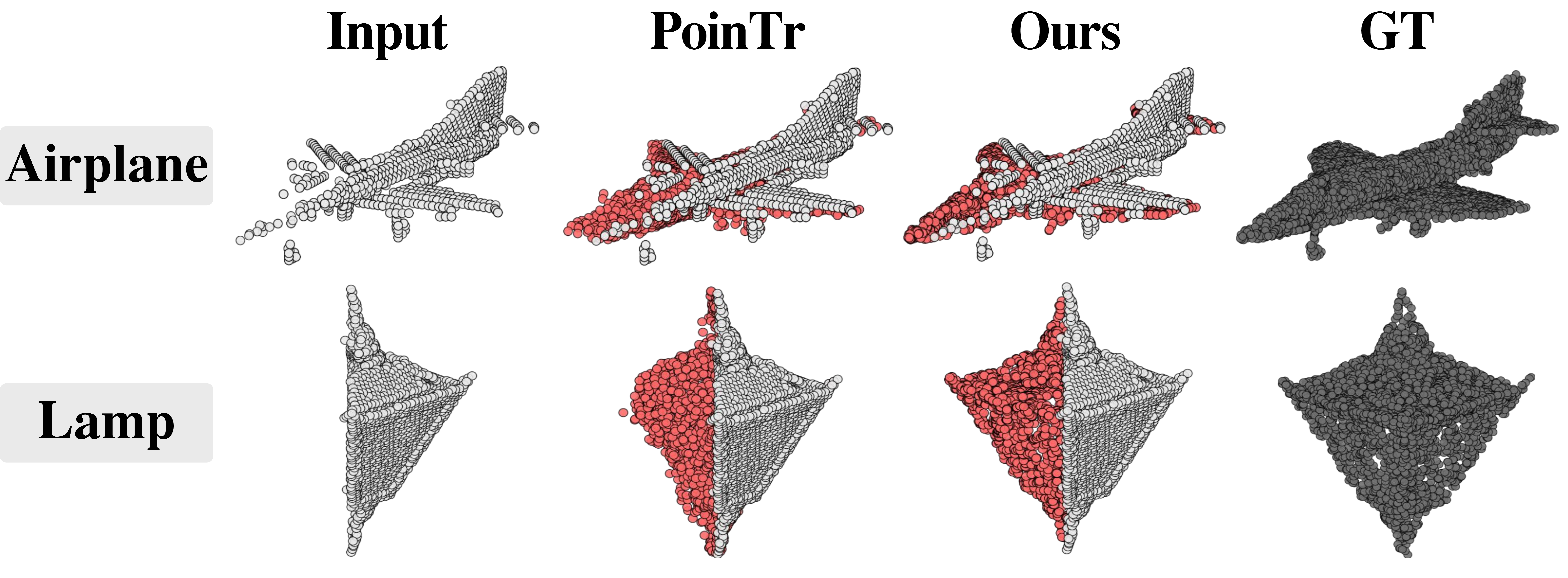}
      \caption{Qualitative comparison with PoinTr and DANCE. Both methods complete only the missing regions, but DANCE achieves more accurate and consistent shapes by leveraging stronger structural and semantic cues.}
      \label{fig:part_result}
   \end{figure}

    This improvement is mainly due to our design, which infers only the missing regions while preserving the input geometry, and leverages semantic guidance from the classification head to provide category-specific shape priors for ambiguous areas.
    To further support this observation, we conducted a qualitative comparison with PoinTr~\cite{yu2021pointr}, a method that also focuses on completing only the missing regions while preserving the input point cloud.
    As shown in Fig.~\ref{fig:part_result}, DANCE achieves more accurate reconstruction of the missing parts, especially in recovering fine-grained geometric structures that PoinTr often fails to capture.

    To assess the robustness of our method under varying densities and challenging conditions, we conducted experiments on the MVP dataset~\cite{pan2021variational}, which provides diverse partial inputs from 26 viewpoints.
    Table~\ref{tab:mvp_comparison} presents comparisons at resolutions of 4,096 and 8,192 points.
    DANCE outperforms existing methods across both density settings, demonstrating robustness to variations in point cloud density.
    We set $R=21$ for the 4,096-point output and $R=33$ for the 8,192-point output to approximately match the target output density.
    These are experimental settings to ensure that the number of candidate points exceeds the desired output size. 
    However, the choice of R is not critical to performance, as discussed in Sec.4.4.
    Since DANCE filters unnecessary candidates based on predicted opacity, the final output naturally approximates the target density. 
    Moreover, in practical scenarios, the output density can be flexibly controlled by adjusting R as needed.

   \begin{table}[t]
    \centering
    \setlength\tabcolsep{5pt}
    \small
    \begin{tabular}{l|cc|cc}
    \hline
    \multirow{2}{*}{Methods} & \multicolumn{2}{c|}{4096} & \multicolumn{2}{c}{8192} \\
    \cline{2-5}
     & CD-Avg $\downarrow$ & F1 $\uparrow$ & CD-Avg$\downarrow$ & F1 $\uparrow$ \\
    \hline
    PCN~\shortcite{yuan2018pcn}   & 7.96 & 0.463 & 6.86 & 0.565\\
    TopNet~\shortcite{tchapmi2019topnet} & 7.69 & 0.434 & 6.64 & 0.526 \\
    ECG~\shortcite{pan2020ecg} & 7.31 & 0.506 & 4.23 & 0.695  \\
    PoinTr~\shortcite{yu2021pointr}  & 4.69 & 0.598 & 3.52 & 0.712  \\
    VRCNet~\shortcite{pan2021variational} & 4.82 & 0.629 & 3.68 & 0.720\\
    SeedFormer~\shortcite{zhou2022seedformer} & 4.61 & 0.605 & 3.59 & 0.710\\
    PDR~\shortcite{lyuconditional}  & 4.42 & 0.638 & 3.45 & 0.739 \\
    DualGenerator~\shortcite{shi2023dualgenerator}  & 4.29 & 0.643 & 3.38 & 0.747 \\
    \hline
    Ours & \textbf{4.19} & \textbf{0.662} & \textbf{3.37} & \textbf{0.754}  \\
    \hline
    \end{tabular}
    \caption{Performance comparison on MVP dataset. DANCE achieves superior results under both 4096 and 8192 input densities, highlighting its robustness to input sparsity.}
    \vspace{-5pt}
    \label{tab:mvp_comparison}
    \end{table}

   \subsection{Real-World Robustness of DANCE}
\begin{figure}[t]
      \centering
      \includegraphics[width=1\linewidth]{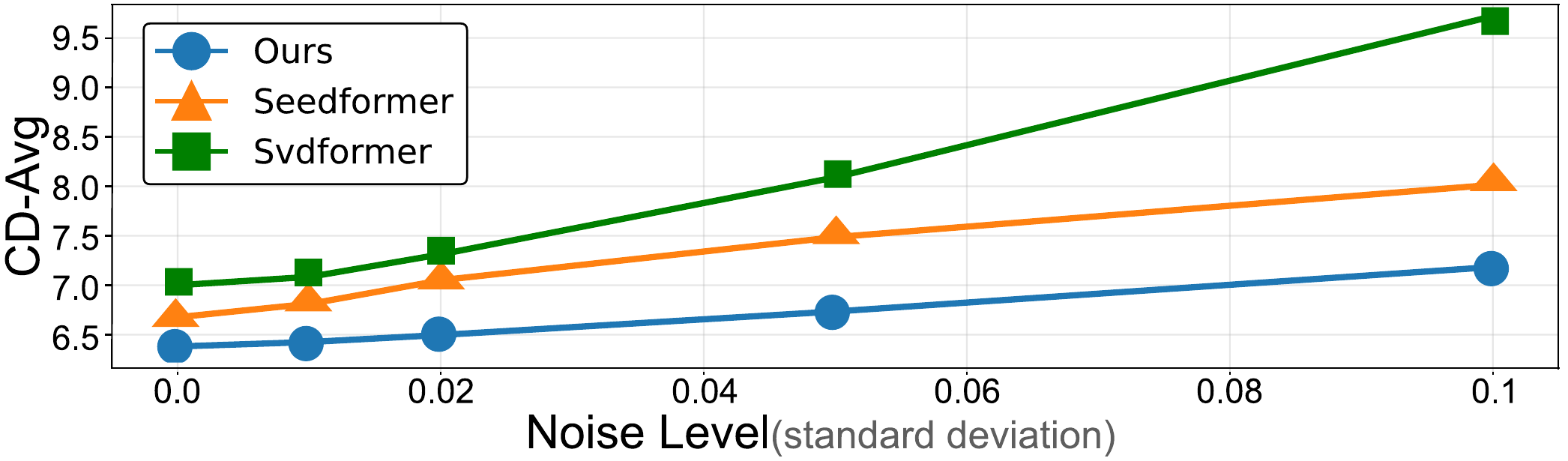}
      \caption{Performance comparison under varying noise levels. DANCE shows strong robustness to noise, exhibiting smaller performance degradation compared to other methods across all noise levels.}
      \label{fig:noise_robust}
   \end{figure}

   We evaluate the robustness of DANCE in two real-world scenarios: (1) noisy inputs from sensor artifacts, and (2) varying input and output point densities.
    Real-world point clouds often contain noise due to sensor limitations or environmental interference.
    To evaluate the robustness of DANCE to such noise, we added Gaussian noise at varying levels to PCN dataset.
    Figure~\ref{fig:noise_robust} shows the change in CD-Avg as the noise level increases.
    DANCE outperforms state-of-the-art methods SVDFormer and SeedFormer, and exhibits greater robustness to various noise levels.

\begin{figure}[t]
      \centering
      \includegraphics[width=0.9\linewidth]{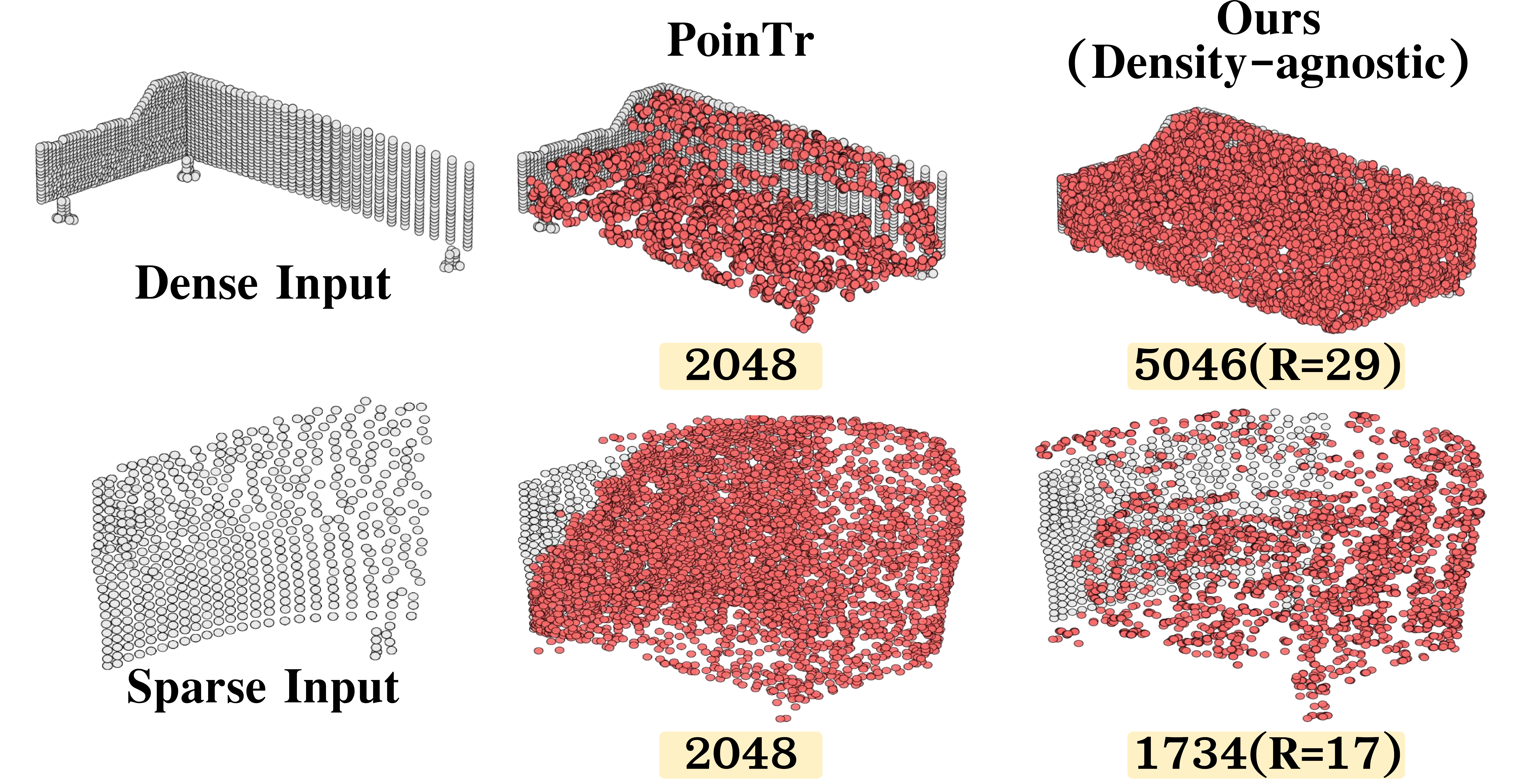}
      \caption{Density-agnostic completion of DANCE. It operates robustly across varying input densities, and the output density is determined according to the sampling parameter $R$ and the predicted opacity values.}
      \vspace{-5pt}
      \label{fig:density_agnostic}
   \end{figure}

   Previous methods typically assume fixed input and output densities, limiting their flexibility in handling varying point cloud sparsity, which does not reflect real-world scenarios.
   In contrast, DANCE achieves density-agnostic completion by leveraging a transformer-based decoder that flexibly operates regardless of the input point cloud density, and adjusts the output density through its opacity prediction process.
   Figure~\ref{fig:density_agnostic} shows that DANCE allows flexible control over the output density. 
   This is achieved by adjusting the face sampling parameter $R$ and filtering out low-opacity candidate points. 
   Although DANCE is trained with a fixed setting of $R=21$, the output density can be modified at inference time by changing $R$ to values like 17 or 29, without requiring retraining. 
   This density-agnostic capability makes DANCE more adaptable to real-world scenarios.
  

   \section{Conclusions and Future Works}
   In this paper, we proposed DANCE, a density-agnostic and class-aware network for point cloud completion.
   DANCE generates candidate points using a ray-based sampling strategy and refines them with a transformer decoder that predicts their position and opacity.
   To provide semantic guidance, we include a simple classification module that learns shape categories directly from incomplete point clouds—without using image-based supervision.
   Thanks to this design, DANCE works well with inputs of different densities and produces flexible, high-quality completions without the need for retraining in different settings.

   However, DANCE currently adopts a fixed sampling configuration, using predetermined viewpoint positions and a constant number of views regardless of the input characteristics.
   While this setting works well in general, it may be suboptimal for objects with highly complex or asymmetric structures.
   As future work, adaptive view sampling strategies can be explored to dynamically adjust the sampling positions and view counts based on the geometric patterns of the input, further improving completion quality in challenging cases.

   \section{Acknowledgment}
   This work was supported by the National Research Foundation of Korea (NRF) grant funded by the Korea government (MSIT) (No. RS-2025-24683045), and by the Institute of Information \& Communications Technology Planning \& Evaluation (IITP) grants funded by the Korea government (MSIT): the Artificial Intelligence Convergence Innovation Human Resources Development (IITP-2023-RS-2023-00256629) and the University ICT Research Center (ITRC) support program (IITP-2025-RS-2024-00437718).

\bibliography{aaai2026}

\end{document}